\documentclass[conference]{IEEEtran}
\IEEEoverridecommandlockouts
\usepackage{cite}
\usepackage{amsmath,amssymb,amsfonts}
\usepackage{algorithmic}
\usepackage{graphicx}
\usepackage{textcomp}
\usepackage{xcolor}
\usepackage{booktabs}
\def\BibTeX{{\rm B\kern-.05em{\sc i\kern-.025em b}\kern-.08em
    T\kern-.1667em\lower.7ex\hbox{E}\kern-.125emX}}
\begin{document}

\title{Few-shot Semantic Encoding and Decoding for Video Surveillance\\
}

\author{\IEEEauthorblockN{1\textsuperscript{st} Baoping Cheng}
\IEEEauthorblockA{\textit{Department of Electronic Engineering} \\
\textit{Tsinghua University}\\
Beijing, China\\
chengbaoping@cmhi.chinamobile.com}
\and
\IEEEauthorblockN{2\textsuperscript{nd} Given Name Surname}
\IEEEauthorblockA{\textit{dept. name of organization (of Aff.)} \\
\textit{name of organization (of Aff.)}\\
City, Country \\
email address or ORCID}
\and
\IEEEauthorblockN{3\textsuperscript{rd} Given Name Surname}
\IEEEauthorblockA{\textit{dept. name of organization (of Aff.)} \\
\textit{name of organization (of Aff.)}\\
City, Country \\
email address or ORCID}
\and
\IEEEauthorblockN{4\textsuperscript{th} Given Name Surname}
\IEEEauthorblockA{\textit{dept. name of organization (of Aff.)} \\
\textit{name of organization (of Aff.)}\\
City, Country \\
email address or ORCID}
\and
\IEEEauthorblockN{5\textsuperscript{th} Given Name Surname}
\IEEEauthorblockA{\textit{dept. name of organization (of Aff.)} \\
\textit{name of organization (of Aff.)}\\
City, Country \\
email address or ORCID}
\and
\IEEEauthorblockN{6\textsuperscript{th} Given Name Surname}
\IEEEauthorblockA{\textit{dept. name of organization (of Aff.)} \\
\textit{name of organization (of Aff.)}\\
City, Country \\
email address or ORCID}
}

\author{\IEEEauthorblockN{Baoping Cheng\IEEEauthorrefmark{1,2},
Yukun Zhang\IEEEauthorrefmark{2}, 
Liming Wang\IEEEauthorrefmark{2},
Xiaoyan Xie\IEEEauthorrefmark{2},
Tao Fu\IEEEauthorrefmark{2},
Dongkun Wang\IEEEauthorrefmark{1},and
Xiaoming Tao\IEEEauthorrefmark{1},}
\IEEEauthorblockA{\IEEEauthorrefmark{1}Department of Electronic Engineering, Tsinghua University, Beijing, China}
\IEEEauthorblockA{\IEEEauthorrefmark{2}China Mobile (Hangzhou) Information Technology Co., Ltd, Hangzhou, China}
\IEEEauthorblockA{Email:\{cbp21, wang-dk23\}@mails.tsinghua.edu.cn, taoxm@tsinghua.edu.cn}
\IEEEauthorblockA{\{zhangyukun, wangliming, xiexiaoyan, futao\}@cmhi.chinamobile.com}
}

\maketitle

\begin{abstract}
With the continuous increase in the number and resolution of video surveillance cameras, the burden of transmitting and storing surveillance video is growing. Traditional communication methods based on Shannon’s theory are facing optimization bottlenecks. Semantic communication, as an emerging communication method, is expected to break through this bottleneck and reduce the storage and transmission consumption of video. Existing semantic decoding methods often require many samples to train the neural network for each scene, which is time-consuming and labor-intensive. In this study, a semantic encoding and decoding method for surveillance video is proposed. First, the sketch was extracted as semantic information, and a sketch compression method was proposed to reduce the bitrate of semantic information. Then, an image translation network was proposed to translate the sketch into a video frame with a reference frame. Finally, a few-shot sketch decoding network was proposed to reconstruct video from sketch. Experimental results showed that the proposed method achieved significantly better video reconstruction performance than baseline methods. The sketch compression method could effectively reduce the storage and transmission consumption of semantic information with little compromise on video quality.  The proposed method provides a novel semantic encoding and decoding method that only needs a few training samples for each surveillance scene, thus improving the practicality of the semantic communication system.
\end{abstract}


\section{Intruduction}
In the past decades, video surveillance has played a crucial role in maintaining the safety and security of society and individuals\cite{b1}. This vital function has led to a significant increase in the number of surveillance cameras and their resolution\cite{b2}. However, the great improvement in video surveillance has also brought a huge burden on video communication and challenges the existing video communication method and system.

Traditional video communication systems build upon the Shannon communication theory, the capacity of which is improved at the cost of system complexity\cite{b3}. The development of these traditional video communication systems has met a bottleneck, and further improvement is facing great difficulties\cite{b4}. Semantic communication, as a new communication paradigm, has received great research attention recently. Semantic communication systems extract and transfer semantic information instead of raw bits and are promising to break the “Shannon trap” and further enhance the capacity of video communication systems\cite{b5}.

There have been several studies on semantic visual communication methods. In 2022, Huang et al. proposed a semantic encoding and decoding method that utilizes category labels, semantic segmentation labels, and features extracted by deep neural networks as semantic information\cite{b6}. In 2023, Li et al. proposed extracting spatio-temporal scene graphs as semantic feature and proposed a spatio-temporal scene graph to video (StSg2vid) model to reconstruct video\cite{b7}. Also in 2023, Du et al. proposed a semantic encoding and decoding method based on sketches\cite{b8}. On the encoding side, the method employs edge detection techniques to obtain the sketch as semantic information. On the decoding side, the method employs the pix2pix\cite{b9} model to reconstruct the original image from the sketch. Experimental results showed that this method can better capture key image features than conventional image compression methods. Although it currently cannot surpass the comparative methods regarding image quality metrics, the research demonstrates the potential of the sketch-based semantic encoding and decoding method in image compression.
\begin{figure*}[htbp]
\centerline{\includegraphics[width=0.9\textwidth]{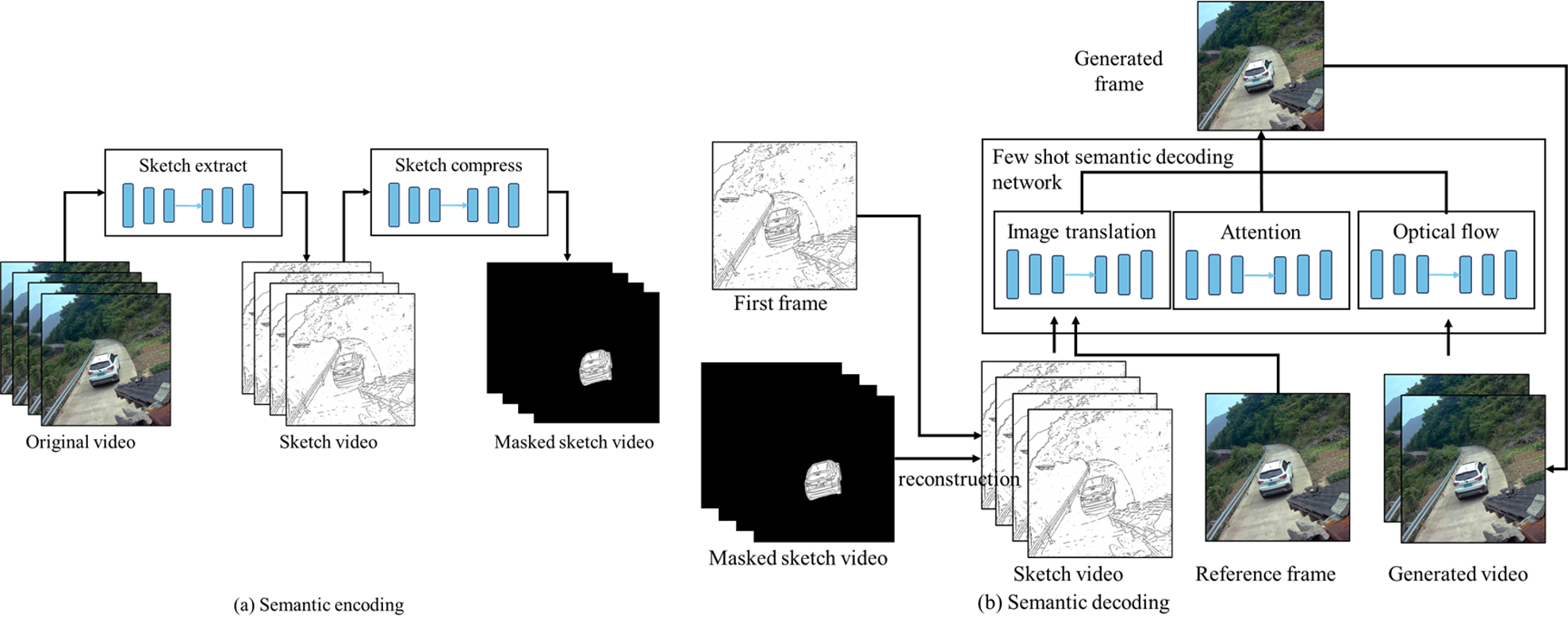}}
\caption{Overall structure of the proposed semantic encoding and decoding method}
\label{fig:ed}
\end{figure*}

In 2024, Du et al. refined their image semantic encoding method based on sketches, extending it to video communication\cite{b10}. This research first extracts edge lines using a deep learning-based edge detection model\cite{b11}. Then it converts the edge detection results into a binary sketch through non-maximum suppression, double-threshold processing, and binarization. Semantic video decoding was achieved through the vid2vid model\cite{b12}. The research results show that this method scores better than traditional video encoding and decoding methods on FID and KID metrics, once again demonstrating the potential of video semantic encoding and decoding methods.

However, existing video semantic encoding and decoding studies generally use deep neural networks such as vid2vid models to reconstruct video. These networks must be trained with many training samples for each video scene. However, collecting and labeling a vast number of surveillance videos is time-consuming and labor-intensive, affecting the practicality of semantic communication. Few-shot learning is a machine learning approach that enables models to learn and make predictions from a small amount of training data\cite{b13}. Developing a few-shot semantic encoding and decoding method is promising in reducing the training samples needed and, therefore, could enhance the practicality of the semantic communication system.

The motivation of this study is to develop a video surveillance semantic encoding and decoding method that only needs a few training samples for each video scene with few-shot learning. The main contributions of this study are listed as follows:

\begin{itemize}
\item A few-shot learning semantic encoding and decoding network based on sketch for video surveillance was proposed. This network only takes a few training samples for each video scene.
\item An image translation neural network that translates a sketch into a video frame was adopted to capture style information from a reference image. enhance the performance of the proposed few-shot semantic decoding network.
\item A static background compression method for sketch video was proposed to reduce the bitrate of video surveillance semantic communication.
\item Throughout, experiments were conducted to evaluate the proposed video surveillance semantic encoding and decoding method. The results show that the proposed methods achieved better surveillance video reconstruction performance than baseline methods. The proposed sketch compression method largely reduces the video storing and transmitting demand.
\end{itemize}

\section{Method}
The proposed semantic encoding and decoding method is illustrated in Fig.~\ref{fig:ed}. For an input video $x = \left\{ {{x_1},{x_2},...,{x_T}} \right\}$, the semantic encoding module extracts the sketch video $s = \left\{ {{s_1},{s_2},...,{s_T}} \right\}$
from the original video and compresses the sketch into a masked sketch video
$ms = \left\{ {m{s_1},m{s_2},...,m{s_T}} \right\}$, where $T$ is the length of the video. The masked sketch video is smaller than the original video. The semantic decoding module first reconstructs the sketch video $\hat s = \left\{ {{{\hat s}_1},{{\hat s}_2},...,{{\hat s}_T}} \right\}$ from the masked sketch video $ms$. Then, generate video frames $\hat x = \left\{ {{{\hat x}_1},{{\hat x}_2},...,{{\hat x}_T}} \right\}$ from the sketch video $\hat s$, reference frame ${x_1}$ and the generated video frames $\left\{ {{{\hat x}_1},{{\hat x}_2},...,{{\hat x}_{t-1}}} \right\}$. The generated video should be similar to the original video.

\subsection{Sketch Extraction and Compression}
\begin{figure}[tb]
\centerline{\includegraphics[width=1.0\linewidth]{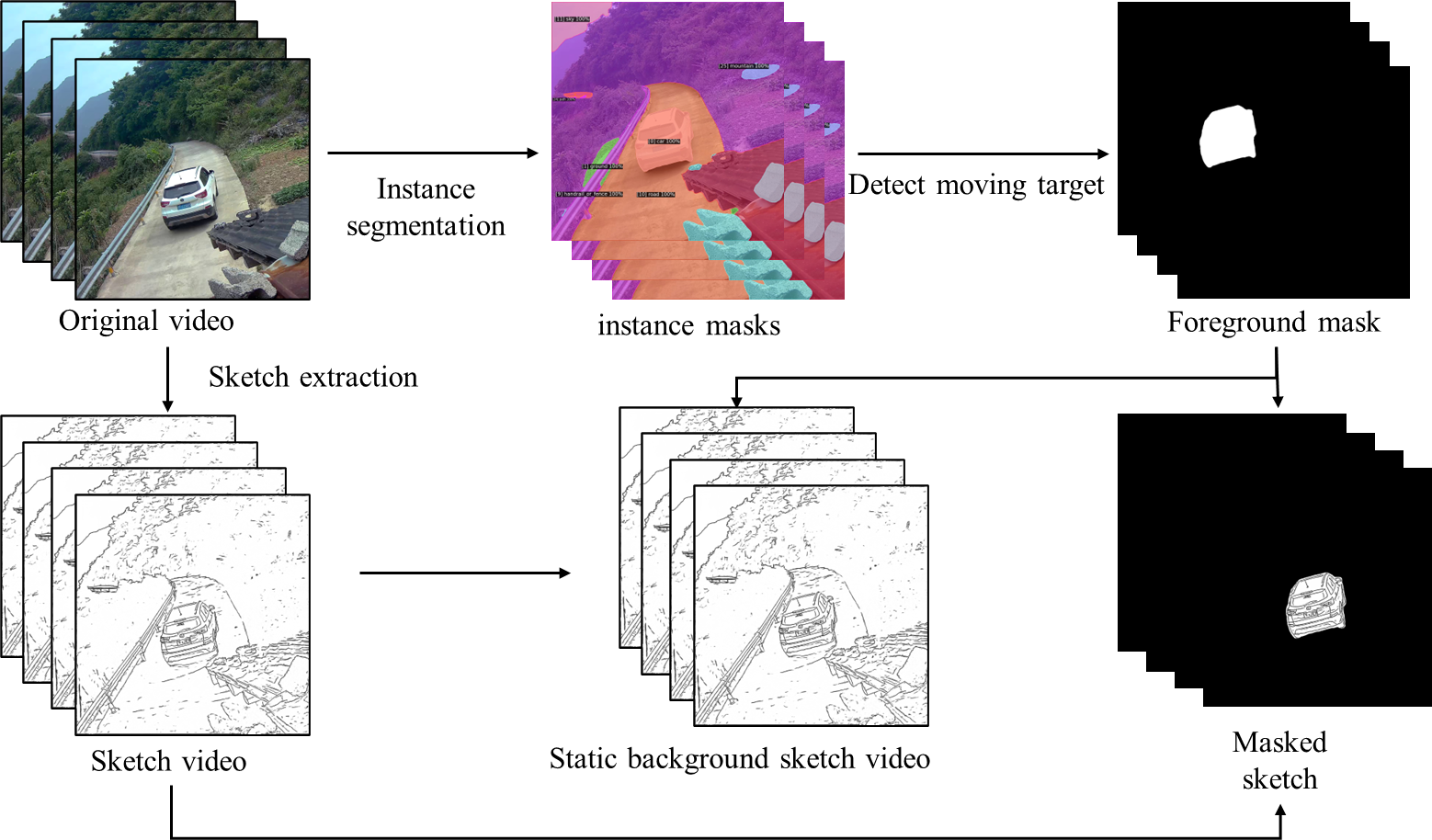}}
\caption{Sketch compression method}
\label{fig:sketchcompression}
\end{figure}

For an original video, a deep learning-based edge detection neural network\cite{b14} was first adopted to extract a sketch video. In surveillance video, the camera is stationary, and the background generally keeps steady. Therefore, there is a lot of redundant semantic information in the background of the sketch video. A sketch compression method (Fig.~\ref{fig:sketchcompression}) was proposed to extract important foregrounds from the sketch video.

A deep learning-based video instance segmentation network\cite{b15} was first adopted to extract segmentation masks from the original video. Multiple instances $I = \left\{ {{I_1},{I_2},...,{I_T}} \right\}$ were detected. For each instance ${I_i}$, there is a mask series $m_1^{{I_i}},m_2^{{I_i}},...,m_T^{{I_i}}$ that indicates the position of the instance in each frame. For each instance, the intersection over union (IoU) was calculated:
\begin{equation}
    Io{U^{{I_i}}} = \frac{{m_1^{{I_i}} \cap m_2^{{I_i}} \cap ,..., \cap m_T^{{I_i}}}}{{m_1^{{I_i}} \cup m_2^{{I_i}} \cup ,..., \cup m_T^{{I_i}}}}
    \label{iou}
\end{equation}

Instances with IoU below a certain threshold were considered as foreground $I_i^F$. Then, masks of all foregrounds is represented as $\{ m_1^{{F_i}},m_2^{{F_i}},...,m_T^{{F_i}}\} _{i = 1}^{{n_F}}$. For each frame, the foreground mask is calculated as follows:
\begin{equation}
    m_t^F = m_t^{F_1^{}} \cup m_t^{F_2^{}} \cup ,..., \cup m_t^{F_{{n_F}}^{}}
    \label{mtf}
\end{equation}

The static background sketch is calculated as follows:
\begin{equation}
    \begin{array}{c}
    s_t^\prime  = m_t^F{s_t} + (1 - m_t^F - (m_1^F \setminus (m_1^F \cap m_t^F))){s_1}\\
    {\rm{       }} + m_1^F \setminus (m_1^F \cap m_t^F){s_T}
    \end{array}
    \label{st}
\end{equation}

To further compress the sketch video, extract the masked sketch as follows:

\begin{equation}
    m{s_t} = m_t^F\max ({s_t},1)
    \label{mst}
\end{equation}

The masked sketch only contains the foreground sketch; pixels of all background parts are 0.

For sketch reconstruction, first calculate the foreground mask:
\begin{equation}
    \hat m_t^F = {\bf{sgn}}(m{s_t})
    \label{mtFhat}
\end{equation}
where ${\bf{sgn}}( \cdot )$ is the sign function. The sketch is then reconstructed as follows:
\begin{equation}
    \begin{array}{c}
    \hat s_t^\prime  = m{s_t} + ({\bf{1}} - \hat m_t^F - (\hat m_1^F \setminus (\hat m_1^F \cap \hat m_t^F))){s_1}\\
    {\rm{      }} + \hat m_1^F{s_T}({\bf{1}} - \hat m_1^F \cap \hat m_t^F)
    \end{array}
    \label{s_t}
\end{equation}

\subsection{The Semantic Decoding Network}
The semantic decoding network was built based on the backbone network vid2vid, which contains an image generation network, a mask prediction network, and an optical flow prediction network. We proposed adding a reference image $y$ input into the network and replacing the image generation network with an image translation network. The proposed network can learn style information from the reference image, thus greatly reducing the training samples for a specific scene needed during network optimization. The overall structure of the semantic decoding network $\cal D$ is 
shown in Fig.~\ref{fig:ed}(b). $\cal D$ is represented as follows:
\begin{equation}
    {\hat x_t} = {\cal D}(\hat x_{t - \tau }^{t - 1},s_{t - \tau }^t,y)
    \label{d}
\end{equation}
The image translation network ${H_{{\theta _H}}}$ takes the sketch of the current frame and the reference image as input and outputs the generated frame ${\hat h_t}$. To improve the quality of video reconstruction by leveraging the correlation between video frames, the semantic decoding network predicts the parts of each video frame that can be mapped from the previously generated image with the optical flow via a mask prediction network. The optical flow prediction network outputs the aforementioned optical flow information. The sketch-based image translation network generates the corresponding part of the image that cannot be obtained through mapping. The networks are represented as follows:
\begin{equation}
    {\hat m_t} = {M_{{\theta _M}}}(\hat x_{t - \tau }^{t - 1},s_{t - \tau }^t)
    \label{mt}
\end{equation}
\begin{equation}
    {\hat w_{t - 1}}{\mkern 1mu}  = {\mkern 1mu} {W_{{\theta _W}}}(\hat x_{t - \tau }^{t - 1},s_{t - \tau }^t)
    \label{wt}
\end{equation}
\begin{equation}
    {\hat h_t} = {H_{{\theta _H}}}({s_t},y)
    \label{ht}
\end{equation}
where ${\theta _M}$, ${\theta _W}$, and ${\theta _H}$ are the learnable parameters.

${\hat w_{t - 1}}$ represents the predicted optical flow between the generated image of the previous frame ${\hat x_{t - 1}}$ and the current frame to be generated ${\hat x_{t}}$. The optical flow prediction network employs a residual network structure. The inputs are the sketches of the past L frames and the current frame $s_{t - \tau }^t = \{ {s_{t - \tau }},{s_{t - \tau  + 1}},...{s_t}\}$ as well as the generated images of the past L frames $\hat x_{t - \tau }^{t - 1} = \{ {\hat x_{t - \tau }},{\hat x_{t - \tau  + 1}},...,{\hat x_{t - 1}}\}$. After obtaining the predicted optical flow, the previous frame’s generated image is warped to obtain the current frame’s warped image 
${\hat w_{t - 1}}\left( {{{\hat x}_{t - 1}}} \right)$

${\hat m_t} = M(\hat x_{t - L}^{t - 1},s_{t - L}^t)$ is the predicted mask, with pixel values ranging from 0 to 1, where 1 indicates complete occlusion with pixels generated by the sketch-based image translation network, and 0 represents complete visibility with pixel values obtained through warping. The combined output is:
\begin{equation}
    {\cal D}(\hat x_{t - \tau }^{t - 1},s_{t - \tau }^t,y) = ({\bf{1}} - {\hat m_t}) \odot {\hat w_{t - 1}}({x_{t - 1}}) + {\hat m_t} \odot {\hat h_t}
    \label{D}
\end{equation}
\begin{figure}[t]
\centerline{\includegraphics[width=1.0\linewidth]{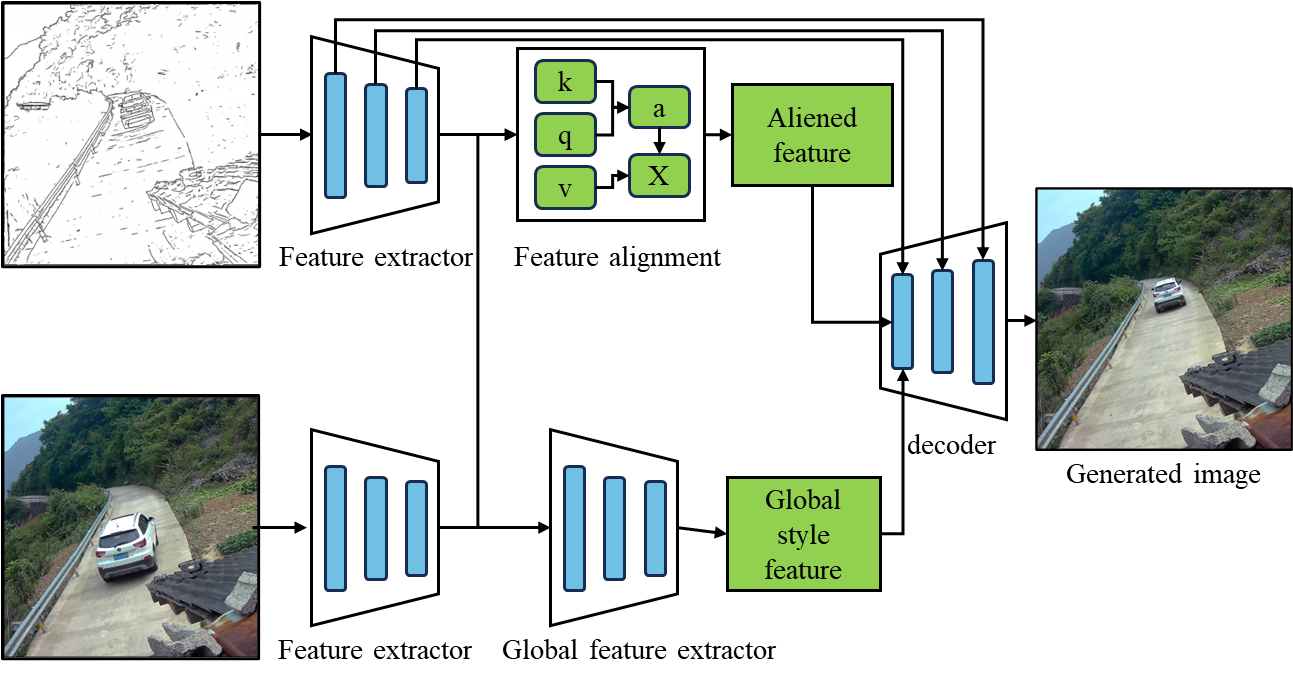}}
\caption{The image translation network}
\label{fig:imagetranslation}
\end{figure}

\subsection{The Image Translation Network}
The overall structure of the proposed image translation network $H$ is shown in Fig.~\ref{fig:imagetranslation}. Given an input sketch ${s_A}$ and a reference image ${y_B}$, where $A$ and $B$ represent the sketch and natural image domains, respectively. The goal of the sketch-based image translation network is to generate an image ${\hat x_B}$ that has a structure similar to the sketch ${s_A}$ and a style similar to the reference image $y_B$. The feature extractors $E_A$ and $E_B$ extract features from the sketch and reference image, respectively. 

The feature alignment network is built upon the MATEBIT backbone \cite{b16}, computing the attention matrix between the sketch and the reference image, and obtaining the aligned features based on the attention matrix. For the sketch features $S_A$ and the reference image features $Y_B$ extracted by the feature extractors $E_A$ and $E_B$, the feature alignment network first computes the query matrix $Q$ from $S_A$ using a $1 \times 1$ convolutional neural network and computes the key matrix $K$  and the value matrix $V$ from $Y_B$. Then, the attention matrix $A^r$ is computed with the query matrix $Q$ and the key matrix $K$: 
\begin{equation}
    {A^r}(u,v) = \frac{{\tilde Q(u)\tilde K{{(v)}^T}}}{{\tilde Q(u) \cdot \tilde K(v)}}
    \label{A}
\end{equation}
where $\widetilde Q(u) = Q(u) - \overline Q (u)$, $\widetilde K(v) = K(v) - \bar K(v)$. $u$ and $v$ are the position index. $\overline Q (u)$ and $\bar K(v)$ are the mean value of $Q(u)$ and $K(v)$. The attention matrix $A^r$ represents the spatial correlation between $S_A$ and $Y_B$.The value matrix is then warped by the attention matrix to acquire the warped image:
\begin{equation}
    {X_{cor}} = {\widetilde A_{mask}}V,\;with\;{\widetilde A_{mask}} = {\rm{softmax}}(\alpha  \cdot {A_{mask}})
    \label{xcor}
\end{equation}
\begin{equation}
    A_{mask}^{} = {\rm{ReLU}}({A^r})
    \label{amask}
\end{equation}
where $\alpha$ controls the sensitivity of the softmax function. The warped image $X_{cor}$ contains the structure information from $S_A$ and the style information from $Y_B$.

The decoder takes the sketch features $S_A$, the warped features$X_{cor}$, and the global style feature $z$ as inputs to output the generated image. The network structure employs the U-Net backbone, with multi-scale sketch features $S_A$ that contain structural information and warped features $X_{cor}$ that contain structural and local style information being incorporated into the various layers of the U-Net via skip connections. The global style feature $z$, which contains global style information, is integrated into the layers of the U-Net using the AdaIN \cite{b17} method. This allows the decoder to acquire structural information from the sketch and fuse the local style information from both the sketch and the reference image, as well as the global style information from the reference image, effectively generating the final output image.

\section{Result and Discussion}

\subsection{Experimental Setting}
A surveillance video dataset is collected from 64 different surveillance cameras, all set in different scenes. For each camera, there are 8 videos, each lasting 2 seconds with a frame rate of 15 frames per second. The resolution of all videos is 512×256. For each camera, 6 videos are randomly selected as training samples, and the remaining 2 videos are test samples. There are, in total, 384 training samples and 128 test samples. As the training sample for each scene is short, the collected dataset challenges the learning ability of the semantic encoding and decoding methods with limited training samples.

Baseline methods include vid2vid\cite{b12}, few-shot vid2vid\cite{b18}, and SGA\cite{b19}. All methods were trained from scratch with the training data. Four generally used metrics are adopted in the experiments:

\begin{itemize}
\item Kernel Inception Distance (KID): KID assesses the quality and diversity of images by comparing the statistics of the features extracted from the generated images to those from a set of real images.
\item Learned Perceptual Image Patch Similarity (LPIPS): LPIPS measures the perceptual similarity between two images, which is closer to human subjective perception.
\item Peak Signal-to-Noise Ratio (PSNR): PSNR measures image quality by comparing the peak signal-to-noise ratio between the original image and the compressed image.
\item Structural Similarity Index (SSIM): SSIM measures the similarity of luminance, contrast, and structure between two images.
\end{itemize}

\subsection{Comparison Experiment}
The video reconstruction results of all compared methods on the KID, LPIPS, PSNR, and SSIM metrics are shown in Tab.~\ref{tab:reconstruct}. “Ours image” and “Ours video” refer to the proposed image translation network and the proposed surveillance video semantic decoding method, respectively. For few-shot vid2vid, SGA, and proposed methods, the first frame of each video was taken as the reference image and was input into the networks. For all methods, videos were reconstructed from the sketch video without compression. For KID and LPIPS, lower values are better. For PSNR and SSIM, higher values are better. 

\begin{table}[tb]
  \centering
  \caption{Video reconstruction comparison}
    \begin{tabular}{p{6.625em}llll}
    \toprule
    \textbf{Methods} & \multicolumn{1}{p{4.19em}}{KID ↓} & \multicolumn{1}{p{4.19em}}{LPIPS↓} & \multicolumn{1}{p{4.19em}}{PSNR ↑} & \multicolumn{1}{p{4.19em}}{SSIM ↑} \\
    \midrule
    vid2vid & 0.2398 & 0.4178 & 18.054 & 0.4753 \\
    fs-vid2vid & 0.2713 & 0.3858 & 18.0359 & 0.5498 \\
    SGA   & 0.3003 & 0.4287 & 18.0661 & 0.4473 \\
    Ours image & 0.2711 & 0.403 & 15.0293 & 0.3802 \\
    Ours video & \textbf{0.1818} & \textbf{0.3164} & \textbf{21.1826} & \textbf{0.6993} \\
    \bottomrule
    \end{tabular}%
  \label{tab:reconstruct}%
\end{table}%

On KID, the proposed video decoding method achieved a 0.1818 KID value, which is 24.2\% lower than the best baseline method vid2vid (0.2398). One-way repeated ANOVA result showed that the decoding method had a significant effect on KID value. Pair t-test results showed that the proposed video decoding method achieved better KID value than all compared methods (t=12.3, 10.5, 12.4, 9.9 respectively, all p<0.01). On LPIPS, the proposed video decoding method achieved a 0.3164 LPIPS value, which is 18.0\% lower than the best baseline method few-shot vid2vid (0.3858). One-way repeated ANOVA result showed that the decoding method had a significant effect on LPIPS value. Pair t-test results showed that the proposed video decoding method achieved better LPIPS value than all compared methods (t=20.3, 16.5, 22.6, 23.8 respectively, all p<0.01). 

Similar results were observed in PSNR and SSIM. the proposed video decoding method achieved best PSNR and SSIM value. One-way repeated ANOVA result showed that the decoding method had a significant effect on PSNR and SSIM values, respectively. Pair t-test results showed that the proposed video decoding method achieved better PSNR and SSIM value than all compared methods respectively(PSNR: t=22.4, 30.6, 18.0, 25.1 respectively, all p<0.01; SSIM: t=36.2, 47.6 39.4, 58.4 respectively, all p<0.01). These results showed that the proposed video decoding method had better decoding performance in the aspect of statistics of the features, perceptual quality, and signal-to-noise ratio. 

The proposed image translation network is part of the proposed video decoding network. The image translation network achieved better performance on KID and LPIPS metrics than compared image generation method SGA (KID: t=3.1, p<0.01; LPIPS: t=6.8, p<0.01), which showed that the proposed image translation network had better perceptual quality. This is because the image translation network took additional reference images and could better capture the style of the video scene. However, the proposed image translation network had worse PSNR and SSIM values (PSNR: t=22.6, p<0.01; SSIM: t=27.4, p<0.01). The proposed video decoding method integrated the image translation network and optical flows and achieved better PSNR and SSIM than SGA. These results showed that the optical flow prediction network and the mask prediction network could enhance the quality of reconstructed video in aspects of structure and signal-to-noise ratio.

This study focused on semantic encoding and decoding; for semantic communication, the sketch and the first frame of each video need to be transmitted. However, semantic transmission is out of the scope of this paper. The proposed method is tested on a two-second video due to the limitation of computation resources. However, the proposed method is able to handle longer videos. 

\subsection{Sketch Compression Experiment}
\begin{figure}[tb]
\centerline{\includegraphics[width=1.0\linewidth]{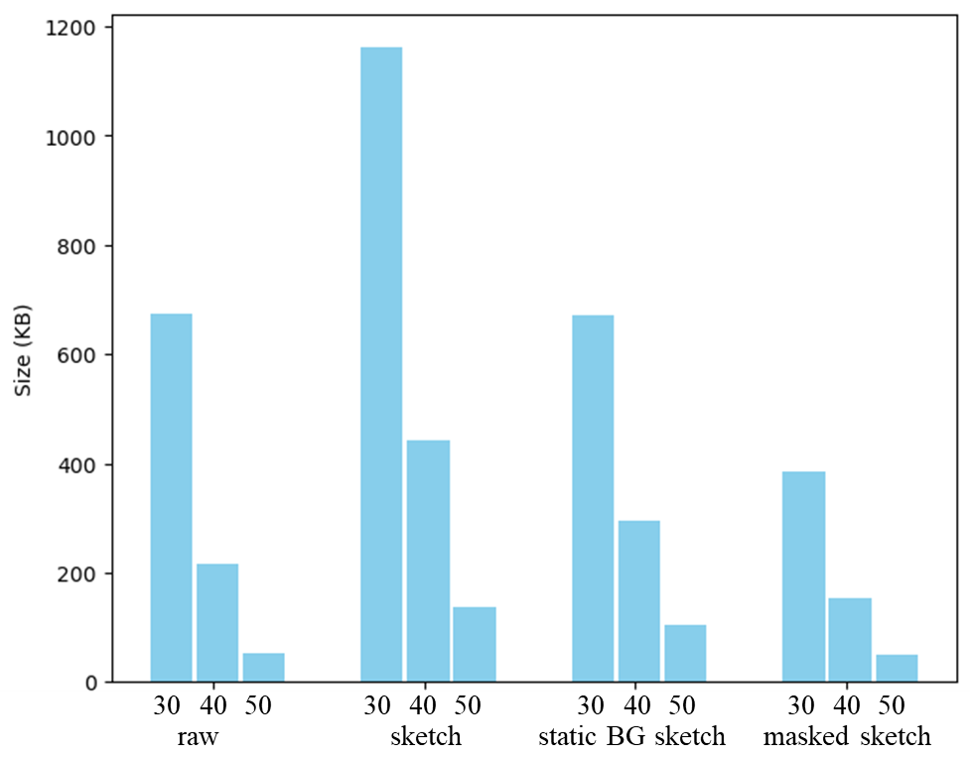}}
\caption{ Sketch compression compare}
\label{fig:sketchcompare}
\end{figure}

Video sizes were compared to evaluate the video compression ability of the proposed sketch compression method. On the self-collected dataset, the raw videos, sketch videos, static background sketch videos, and masked sketch videos were encoded by H.264 with Quantization Parameter (QP) equals 30, 40, and 50 separately. The corresponding results are shown in Tab.~\ref{fig:sketchcompare}. The size of raw videos under QP 30, 40, and 50 are 674±337 KB, 215±113 KB, and 53±26 separately. The size of masked sketch videos under QP 30, 40, and 50 are 384±343 KB, 153±134 KB, and 49±42 separately. Two-way Analysis of Variance (ANOVA) result showed that both video type and QP had a significant effect on video size (F=465, 4688 separately, all p<0.01). Besides, the interaction between video type and QP was significant (F=383, p<0.01). Paired t-test result showed that the masked sketch had a significantly smaller video size than the raw video and sketch video under all QP values (masked sketch vs. raw: t=15.2, 9.4, 2.3, p<0.01, 0.01, 0.05, respectively; masked sketch vs. sketch: t=28.0, 28.0, 28.3 respectively, all p<0.01). These results showed that the proposed sketch compression method could significantly reduce the storage and transfer bit rate of sketch video and achieve a lower bit rate than raw video. Besides, the sketch compression ratio is higher with a lower QP value, which means that the proposed sketch compression method works better with high video quality. 

The matrices of the video reconstructed from the original sketch and masked sketch are shown in Tab.~\ref{fig:sketchcompare}. The paired t-test result showed that the video reconstructed with the original sketch has better performance on KID, LPIPS, and PSNR metrics (t=3.7, 3.4, 4.9 respectively, all p<0.01) but not significant on SSIM (t=1.1, p=0.27). Although the masked sketch compromised performance on most of the metrics, the performance difference is rather small. The metrics values of the masked sketch were 2.9\%, 0.5\%, 0.8\%, and 0.1\% worse than the original sketch. Besides, the reconstructed video with masked sketch still achieved better performance than the video reconstructed from the original sketch with all compared methods (all p<0.01). The above result showed that the proposed sketch compression method could effectively reduce the bitrate of encoded video with little performance decline.

\begin{table}[tb]
  \centering
  \caption{Video reconstruction comparison with different sketch}
    \begin{tabular}{p{6.625em}llll}
    \toprule
    \textbf{Methods} & \multicolumn{1}{p{4.19em}}{KID ↓} & \multicolumn{1}{p{4.19em}}{LPIPS↓} & \multicolumn{1}{p{4.19em}}{PSNR ↑} & \multicolumn{1}{p{4.19em}}{SSIM ↑} \\
    \midrule
    Masked sketch & 0.187 & 0.3181 & 21.0203 & 0.6989 \\
    Original sketch & \textbf{0.1818} & \textbf{0.3164} & \textbf{21.1826} & \textbf{0.6993} \\
    \bottomrule
    \end{tabular}%
  \label{tab:sketch compare}%
\end{table}%

\begin{figure*}[t]
\centerline{\includegraphics[width=0.9\textwidth]{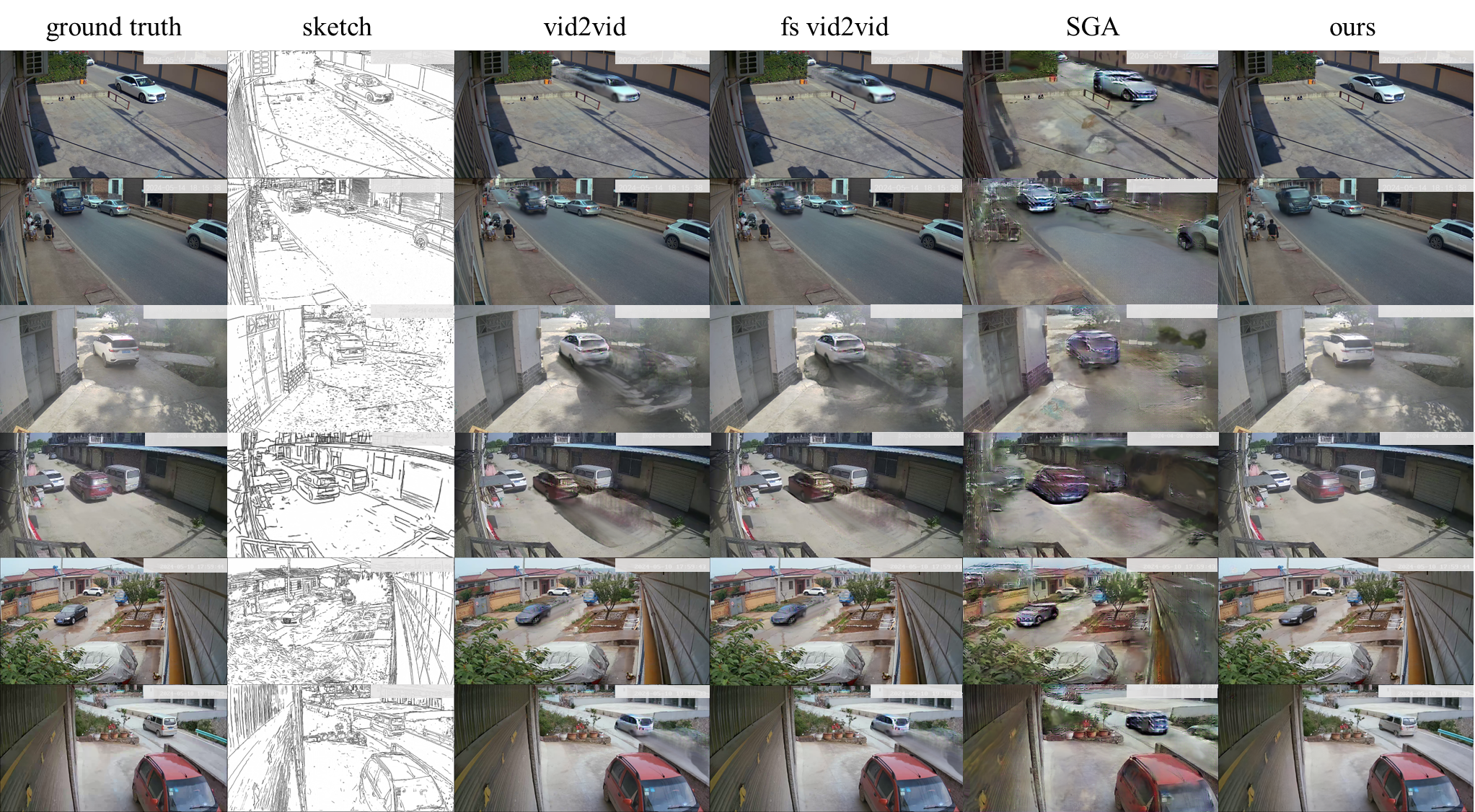}}
\caption{Reconstructed frames of different video decoding method}
\label{fig:visualization}
\end{figure*}

\subsection{Visualization results}

The reconstructed video frames are shown in Fig.~\ref{fig:visualization}. For human perception, SGA has the worst performance, both the foreground and the background had obvious blur and distortion. For the first and fourth rows, obvious motion blur is shown in vid2vid and few-shot vid2vid, while the proposed method had little such problem. In other examples, all methods failed to reconstruct exact details, such as license plates and wheels. When zoomed in, the reconstructed foreground of the proposed method showed different degrees of blur and distortion. However, the proposed method showed a better ability to capture the general color and structure of foreground targets than baseline methods and exhibited a better overall visual appearance.


\section{Conclusion}

In this study, a semantic encoding and decoding method for surveillance video was proposed. The sketch was taken as semantic information, and a sketch compression method was proposed to reduce the storing and transmitting consumption of the sketch. A few-shot video decoding network was proposed, which only needs a few samples for each surveillance scene to be trained. An image translation network was adopted to the video decoding network that captures style information from a reference image. Experimental results indicated that the proposed method can achieve lower storing and transmitting consumption and better video reconstruction performance than baseline methods with only six training samples for each surveillance scene. The proposed method enhances the practicality of the surveillance video semantic communication systems.


\end{document}